\documentclass[sigconf]{acmart}
\usepackage{array}
\usepackage{booktabs}
\usepackage{multirow}
\usepackage{subcaption}
\usepackage{graphicx}
\usepackage{adjustbox}

\newcolumntype{P}[1]{>{\raggedright\arraybackslash}p{#1}}
\usepackage{subcaption}
\usepackage{listings}
\usepackage{xcolor}
\usepackage{multirow}
\definecolor{codegreen}{rgb}{0,0.6,0}
\definecolor{codegray}{rgb}{0.5,0.5,0.5}
\definecolor{codepurple}{rgb}{0.58,0,0.82}
\definecolor{backcolour}{rgb}{0.95,0.95,0.92}

\lstdefinestyle{mystyle}{
    backgroundcolor=\color{backcolour},   
    commentstyle=\color{codegreen},
    keywordstyle=\color{magenta},
    numberstyle=\tiny\color{codegray},
    stringstyle=\color{codepurple},
    basicstyle=\ttfamily\footnotesize,
    breakatwhitespace=false,         
    breaklines=true,                 
    captionpos=b,                    
    keepspaces=true,                 
    numbers=left,                    
    numbersep=5pt,                  
    showspaces=false,                
    showstringspaces=false,
    showtabs=false,                  
    tabsize=2
}

\AtBeginDocument{%
  }

\setcopyright{acmlicensed}
\copyrightyear{2018}
\acmYear{2018}
\acmDOI{XXXXXXX.XXXXXXX}
\acmConference[Conference acronym 'XX]{Make sure to enter the correct
  conference title from your rights confirmation email}{June 03--05,
  2018}{Woodstock, NY}
\acmISBN{978-1-4503-XXXX-X/2018/06}

\begin{document}

\title{\texttt{MissMecha}: An All-in-One Python Package for Studying Missing Data Mechanisms}

\author{Youran Zhou}
\affiliation{%
  \institution{Deakin University}
  \city{Geelong}
  \country{Australia}}
  \email{echo.zhou@deakin.edu.au}

\author{Mohamed Reda Bouadjenek}
\affiliation{%
  \institution{Deakin University}
  \city{Geelong}
  \country{Australia}}
\email{reda.bouadjenek@deakin.edu.au}

\author{Sunil Aryal}
\affiliation{%
  \institution{Deakin University}
  \city{Geelong}
  \country{Australia}}
\email{sunil.aryal@deakin.edu.au}





\renewcommand{\shortauthors}{Zhou et al.}

\begin{abstract}
Incomplete data is a persistent challenge in real-world datasets, often governed by complex and unobservable missing mechanisms. Simulating missingness has become a standard approach for understanding its impact on learning and analysis. However, existing tools are fragmented, mechanism-limited, and typically focus only on numerical variables, overlooking the heterogeneous nature of real-world tabular data. We present \texttt{MissMecha}, an open-source Python toolkit for simulating, visualizing, and evaluating missing data under MCAR, MAR, and MNAR assumptions. MissMecha supports both numerical and categorical features, enabling mechanism-aware studies across mixed-type tabular datasets. It includes visual diagnostics, MCAR testing utilities, and type-aware imputation evaluation metrics. Designed to support data quality research, benchmarking, and education, \texttt{MissMecha} offers a unified platform for researchers and practitioners working with incomplete data. Documentation and interactive notebooks are available at :
\href{https://echoid.github.io/MissMecha/index.html}{\texttt{echoid.github.io/MissMecha}}. 

\noindent Demo video: \url{https://vimeo.com/1094349687/40d4c218e7}
\end{abstract}




\maketitle

\section{Introduction}
\textbf{Motivation.}  
Missing data is a pervasive challenge in applied machine learning, particularly in domains such as healthcare, finance, and the social sciences~\cite{healthcareDL,miao2022experimental,zhou2024reviewhandlingmissingdata,mldl,iot,zhou2025developingrobustmethodshandle,DDIM,hipmk}. Incomplete datasets introduce uncertainty into modeling pipelines, often leading to biased estimates and reduced generalizability. While a wide range of imputation methods have been proposed, the assumptions underlying why data is missing—known as missingness mechanisms~\cite{rubin1976inference,rubin2004multiple,little}—are rarely examined or tested explicitly. Simulation has therefore become a standard approach: researchers inject controlled missing values into complete datasets to benchmark model robustness. 

However, existing simulation practices suffer from key limitations. (1) many studies manually generate missingness with inconsistent ratios and heuristics, resulting in incomparable setups and a lack of standardized benchmarks—ultimately slowing progress in incomplete data research.~\cite{jager2021benchmark,Subtypesofthemissing} (2) , most available tools and assumptions focus solely on numerical variables, overlooking the heterogeneous nature of real-world tabular data that contains both categorical and continuous attributes. This restricts the diversity of test cases and fails to fully assess an imputation model’s utility. (3) , when simulating heterogeneous missingness, metric adaptation becomes necessary: categorical variables require different error formulations and baselines to ensure fair evaluation, yet few tools support such extensions. (4), although missing mechanism detection remains an open problem, structure-aware visualizations and statistical tests (e.g., MCAR diagnostics)~\cite{missmech, alabadla2022systematic} can provide useful signals for analysts, aiding downstream imputation choices and mechanism hypothesis validation.

\textbf{Related Works.}
Although missingness mechanisms critically affect the validity of imputation, most existing studies use ad-hoc and non-reproducible simulation setups, varying in rates, patterns, and assumptions. Tools to standardize this process remain limited. \texttt{pyampute}~\cite{pyampute} supports flexible MCAR/MAR simulation but lacks built-in evaluation or mixed-type support. \texttt{OT}~\cite{OT} includes hardcoded MCAR logic but no mechanism flexibility. R’s \texttt{missMethods}~\cite{missMethods} and MATLAB’s \texttt{SMD}~\cite{synthetic-missing-data} offer statistical methods without extensibility or evaluation tools. \texttt{missingno}~\cite{pyampute} focuses on visualization; \texttt{MissMech}~\cite{missmech} supports MCAR testing but lacks simulation or pipeline integration. As shown in Table~\ref{tab:toolkit-comparison}, no prior toolkit supports heterogeneous tabular data in an integrated, extensible way. \texttt{MissMecha} addresses these gaps by combining mechanism-aware simulation, visualization, statistical testing, and evaluation in a unified Python framework.
 
\textbf{Our Contributions.}
We introduce \texttt{MissMecha}, an open-source Python toolkit for mechanism-aware simulation, visualization, and evaluation of missing data in heterogeneous tabular datasets. It supports MCAR, MAR, and MNAR mechanisms across numerical and categorical variables, with a modular design:

The \texttt{generate} module provides over a dozen missingness strategies (e.g., logistic, correlation-, or quantile-based), supporting both global and column-wise control via a scikit-learn-style API. The \texttt{visual} module offers heatmaps, correlation plots, and bar charts in customizable styles for diagnostics and teaching. The \texttt{analysis} module includes Little’s MCAR test, missingness summaries, and type-aware imputation evaluation via standard metrics and \texttt{AvgErr}. The \texttt{impute} module provides \texttt{SimpleSmartImputer}, a baseline imputer that auto-selects mean or mode by type.

\texttt{MissMecha} unifies simulation and evaluation in a reproducible, extensible framework—ideal for research, benchmarking, and education. Available at \href{https://pypi.org/project/missmecha-py/}{\texttt{missmecha-py}}.


\begin{table}[ht]
\caption{
Feature Comparison of Missing Data Toolkits.
\textbf{Mech.}: Supports missingness generation (MCAR/MAR/MNAR);
\textbf{Eval.}: Imputation evaluation metrics;
\textbf{Tests}: Statistical mechanism testing (e.g., MCAR test);
\textbf{Vis.}: Visualization of missing patterns;
\textbf{Impute}: Built-in or compatible imputation interface;
\textbf{Hetero.}: Supports mixed-type (categorical + numerical) data.
}
\label{tab:toolkit-comparison}
\centering
\setlength{\tabcolsep}{4pt}
\begin{tabular}{lccccccc}
\toprule
\textbf{Tool} & \textbf{Lang.} & \textbf{Mech.} & \textbf{Eval.} & \textbf{Tests} & \textbf{Vis.} & \textbf{Imp} & \textbf{Hetero.} \\
\midrule
\cite{pyampute}   & Py &\checkmark &  $\times$ &\checkmark &  $\times$ &  $\times$    &  $\times$ \\
\cite{missingno} & Py &  $\times$ &  $\times$ &  $\times$ &\checkmark &  $\times$    &  $\times$ \\
\cite{OT}            & Py &\checkmark &\checkmark &  $\times$ &  $\times$ &\checkmark     &  $\times$ \\
\cite{missMethods} & R &\checkmark &  $\times$ &\checkmark &  $\times$ &\checkmark     &  $\times$ \\
\cite{missmech}     & R &  $\times$ &  $\times$ &\checkmark &  $\times$ & Base  &  $\times$ \\
\cite{synthetic-missing-data}          & Matlab &\checkmark &  $\times$ &  $\times$ &  $\times$ &  $\times$     &  $\times$ \\\midrule\midrule
Ours     & Py &\checkmark &\checkmark &\checkmark &\checkmark & Base  &\checkmark \\
\bottomrule
\end{tabular}
\end{table}

\section{MissMecha Framework}
Figure~\ref{fig:architecture} illustrates the architecture of \texttt{MissMecha}, a modular and extensible Python toolkit for simulating and analyzing missing data mechanisms in tabular datasets. It unifies missingness generation, visualization, statistical testing, and evaluation within a consistent interface, supporting reproducible experimentation, benchmarking, and educational use. \texttt{MissMecha} is designed for researchers studying missing data, data scientists evaluating imputation robustness, and instructors teaching structured missingness. It is particularly valuable in domains such as healthcare, finance, and social science, where incomplete data is pervasive. The toolkit includes four main modules: a flexible missingness generator, a visual diagnostics module, an analysis module for mechanism testing and imputation evaluation, and a baseline imputer supporting heterogeneous data. All components follow a \texttt{scikit-learn}-style interface for seamless integration into existing workflows. Documentation and interactive notebooks are available at: \href{https://echoid.github.io/MissMecha/}{\texttt{echoid.github.io/MissMecha/}}. Demo video: \url{https://vimeo.com/1094349687/40d4c218e7}

\begin{figure}[h]
  \centering
  \includegraphics[width=\linewidth]{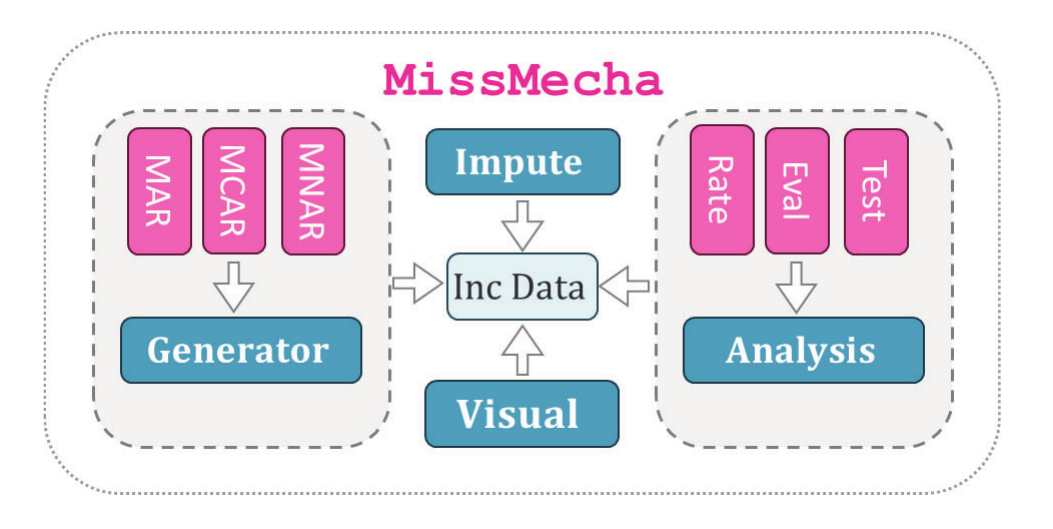}
  \caption{
  Architecture of \texttt{MissMecha}, comprising four modules: a configurable missingness generator (MCAR, MAR, MNAR), a visual diagnostics module, an analysis module for statistical testing and evaluation, and a baseline imputer. All modules follow a unified API and support heterogeneous tabular data.
  }
  \label{fig:architecture}
\end{figure}

\section{MissMecha Component}


\subsection{Generate Module}

At the core of \texttt{MissMecha} is the \texttt{generate} module, which provides a flexible and extensible framework for simulating missing values under different assumptions. It supports MCAR, MAR, and MNAR mechanisms, each implemented with multiple strategy variants. Missingness can be introduced globally or on a per-column basis, and supports both numerical and categorical features, as well as mixed-type tabular data, via a unified dictionary-based configuration.

The main interface class, \texttt{MissMechaGenerator}, follows a \texttt{scikit-learn}-style \texttt{fit}/\texttt{transform} API, enabling users to learn mechanism-specific parameters from one dataset and apply the same configuration to others with a compatible schema. The typical usage pattern includes:

\begin{itemize}
  \item \texttt{fit(X, y=None)} – prepares internal parameters based on the input data; for label-aware mechanisms (e.g., MARType2, MNARType2), \texttt{y} can be provided to guide simulation using class-conditional information;
  \item \texttt{transform(X)} – applies the configured missingness mechanism to the input data;
  \item \texttt{fit\_transform(X, y=None)} – performs fitting and transformation in a single step;
  \item \texttt{get\_mask()} and \texttt{get\_bool\_mask()} – return binary or boolean masks indicating observed and missing entries.
\end{itemize}

Users can customize key parameters such as the mechanism type (e.g., logistic masking, self-censoring), the overall missing rate, transformation functions, and dependency columns for structured dropout. Categorical variables are handled automatically through ordinal encoding and threshold-based logic, enabling seamless simulation across heterogeneous datasets without additional preprocessing. Currently, the module supports 3 MCAR types, 8 MAR types, and 6 MNAR types. Mechanism details and their behaviors are summarized in Table~\ref{tab:mechanism_summary}.

\begin{lstlisting}[language=Python]
from missmecha.generator import MissMechaGenerator

# Initialize the generator with MCAR Type 1 and 20% missing rate
generator = MissMechaGenerator(mechanism="mcar", mechanism_type=1, missing_rate=0.2)

# Apply simulation
X_missing = generator.fit_transform(data)
\end{lstlisting}

\textbf{Listing 1:} Example usage of the \texttt{MissMechaGenerator} class to simulate MCAR missingness using Type 1 masking. The interface adopts a consistent \texttt{fit}/\texttt{transform} workflow compatible with \texttt{scikit-learn}.

\subsubsection{Mechanism Functions}

The currently supported missing data mechanisms in \texttt{MissMecha} are summarized in Table~\ref{tab:mechanism_summary}. Each mechanism is implemented as a separate class under the \texttt{missmecha.generate} subpackage, following a standardized interface. Examples include \texttt{MCARType1}, \texttt{MARType4}, and \texttt{MNARType3}. Each mechanism class accepts a set of configuration parameters tailored to the simulation strategy it implements. While core parameters such as \texttt{missing\_rate} and \texttt{seed} are shared across all types, additional parameters (e.g., \texttt{up\_percentile}, \texttt{obs\_percentile}, or \texttt{depend\_on}) are specific to certain mechanisms. This reflects the diversity of internal assumptions and masking behaviors among different missingness models. A full list of supported parameters and their usage is provided in the official documentation. Table~\ref{tab:mechanism_summary} summarizes all currently available mechanism types. Additional mechanisms will be introduced in future versions of the toolkit. Importantly, \texttt{MissMecha} is designed to be easily extensible. Users can define custom mechanisms by subclassing the base structure of existing implementations. Any new class that follows the standard \texttt{fit}/\texttt{transform} interface and produces a compatible output shape can be directly integrated into the \texttt{MissMechaGenerator} for use in downstream simulation pipelines.

\begin{table}[ht]
\caption{Summary of missing data mechanisms implemented in \texttt{MissMecha}.}
\label{tab:mechanism_summary}
\centering
\footnotesize  
\setlength{\tabcolsep}{3pt}
\begin{tabular}{@{}llcP{4.5cm}@{}}
\toprule
\textbf{Mech.} & \textbf{Name} & \textbf{Type} & \textbf{Description} \\
\midrule
\multirow{3}{*}{MCAR}
& Uniform masking & 1 & Each cell has an equal independent chance of being missing. \\
& Fixed selection & 2 & Randomly masks a fixed number of cells across the dataset. \\
& Column-balanced & 3 & Applies missingness evenly across all columns. \\
\midrule
\multirow{8}{*}{MAR}
& Logistic model & 1 & Missingness depends on features via logistic regression. \\
& Mutual information & 2 & Columns with high MI to label drive masking. \\
& Point-biserial & 3 & Uses label correlation to control masking. \\
& Correlation ranking & 4 & Masks columns based on pairwise correlations. \\
& Rank-based masking & 5 & Uses ranked values in control column to assign masking. \\
& Binary grouping & 6 & Splits rows by median; applies uneven masking. \\
& Top-value rule & 7 & Keeps rows with top values and masks others. \\
& Extreme-value & 8 & Masks rows with high and low extremes. \\
\midrule
\multirow{6}{*}{MNAR}
& Quantile thresholding & 1 & Masks values above/below quantile cutoffs. \\
& Logistic self-dependence & 2 & Missingness depends on observed values in same row. \\
& Self-masking & 3 & Feature masks itself based on its own value. \\
& Quantile cut & 4 & Applies upper/lower/both quantile-based cuts. \\
& Feature-wise masking & 5 & Self-masking applied independently per column. \\
& Percentile masking & 6 & Masks values below percentile per column. \\
\bottomrule
\end{tabular}
\end{table}

\subsubsection{Column-wise Parameters}

In addition to global missingness simulation, \texttt{MissMechaGenerator} also supports fine-grained column-wise control via the \texttt{info} parameter. This allows users to specify different mechanisms, types, and missing rates for individual columns or groups of columns, enabling more realistic and heterogeneous simulation setups. Each entry in the \texttt{info} dictionary defines a per-column configuration, including mechanism type (e.g., MCAR, MAR, MNAR), its variant, missing rate, and optional dependencies (such as other feature names or additional parameters). 

\subsection{Analysis Module}

The \texttt{analysis} module supports systematic exploration of missingness structure and imputation quality through three key components:

\subsubsection{Missingness summaries.}  
The function \texttt{compute\_missing\_rate} reports column-wise and overall missing rates in a dataset. It supports both \texttt{pandas.DataFrame} and \texttt{numpy.ndarray} inputs, and optionally prints a formatted summary and visualizes a bar chart of missingness per feature. This is useful for understanding the coverage profile of simulated or real-world data.

\subsubsection{Imputation evaluation.}  
To assess the quality of imputed values, \texttt{evaluate\_imputation} compares an imputed dataset to the original ground truth at previously missing positions. The function supports mixed-type data by applying RMSE or MAE for numerical columns and accuracy for categorical columns. Users can provide a list of categorical column names to enable type-aware evaluation, and results are returned as both raw and normalized (0–1 scaled) scores to mitigate the effect of differing feature scales. \texttt{MissMecha} also introduces \texttt{AvgErr}~\cite{hivae}, a hybrid metric for unified imputation assessment across heterogeneous datasets, combining scaled numeric error with categorical accuracy.

\subsubsection{Statistical testing of missingness mechanisms.}
The class \texttt{MCARTest} provides formal hypothesis tests to assess whether the missingness mechanism aligns with the MCAR assumption. It includes a global test via \texttt{little\_mcar\_test}, which computes Little’s MCAR statistic and returns a single p-value, and \texttt{mcar\_t\_tests}, which performs pairwise t-tests between missing and observed groups for each feature, yielding a matrix of p-values. An auxiliary method \texttt{report} is also provided to help interpret and summarize the test results.

\begin{figure*}[ht]
  \centering
  \begin{subfigure}{0.32\textwidth}
    \includegraphics[height=4cm,width=\linewidth]{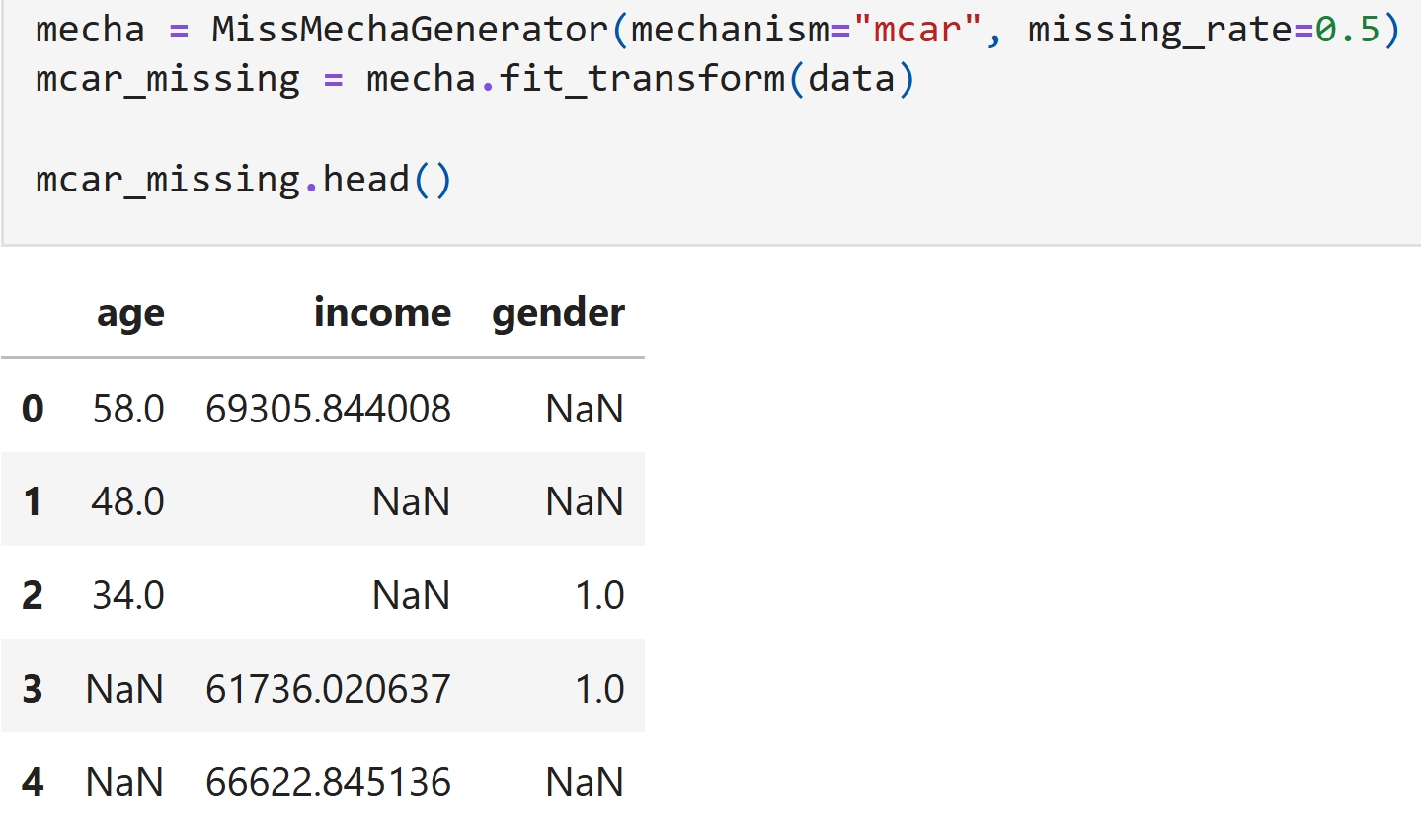}
    \caption{MCAR simulation.}
  \end{subfigure}
  \hfill
  \begin{subfigure}{0.32\textwidth}
    \includegraphics[height=4cm,width=\linewidth]{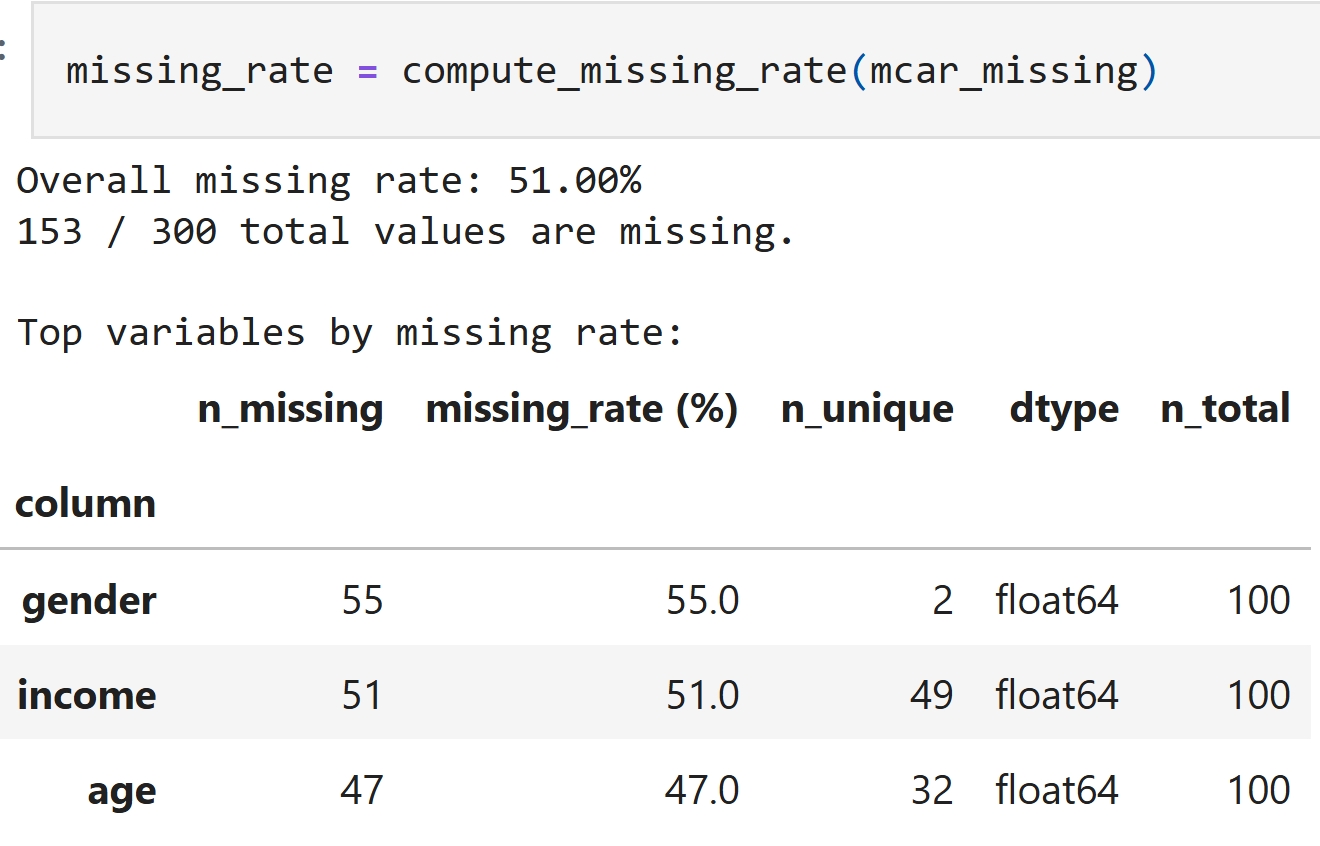}
    \caption{Missing rate summary.}
  \end{subfigure}
  \hfill
  \begin{subfigure}{0.32\textwidth}
    \includegraphics[width=\linewidth]{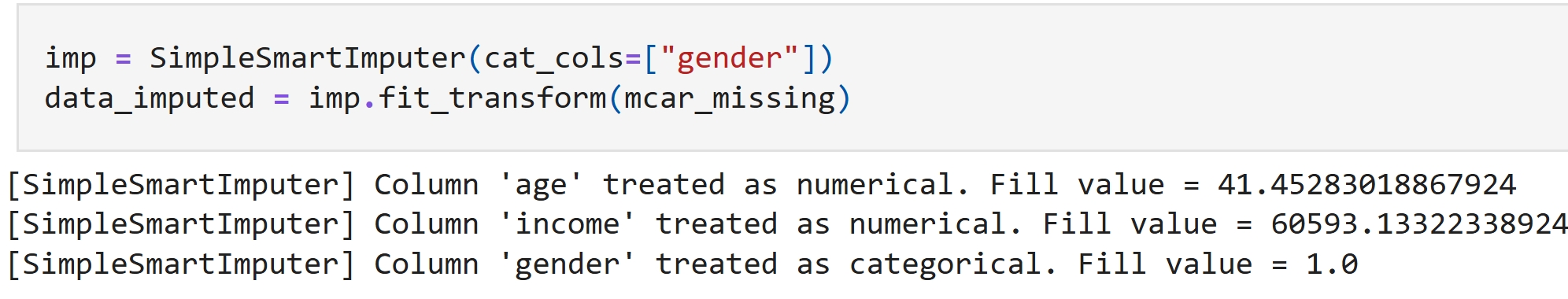}
    \caption{Imputation configuration.}
  \end{subfigure}

  \begin{subfigure}{0.32\textwidth}
    \includegraphics[height=4cm,width=\linewidth]{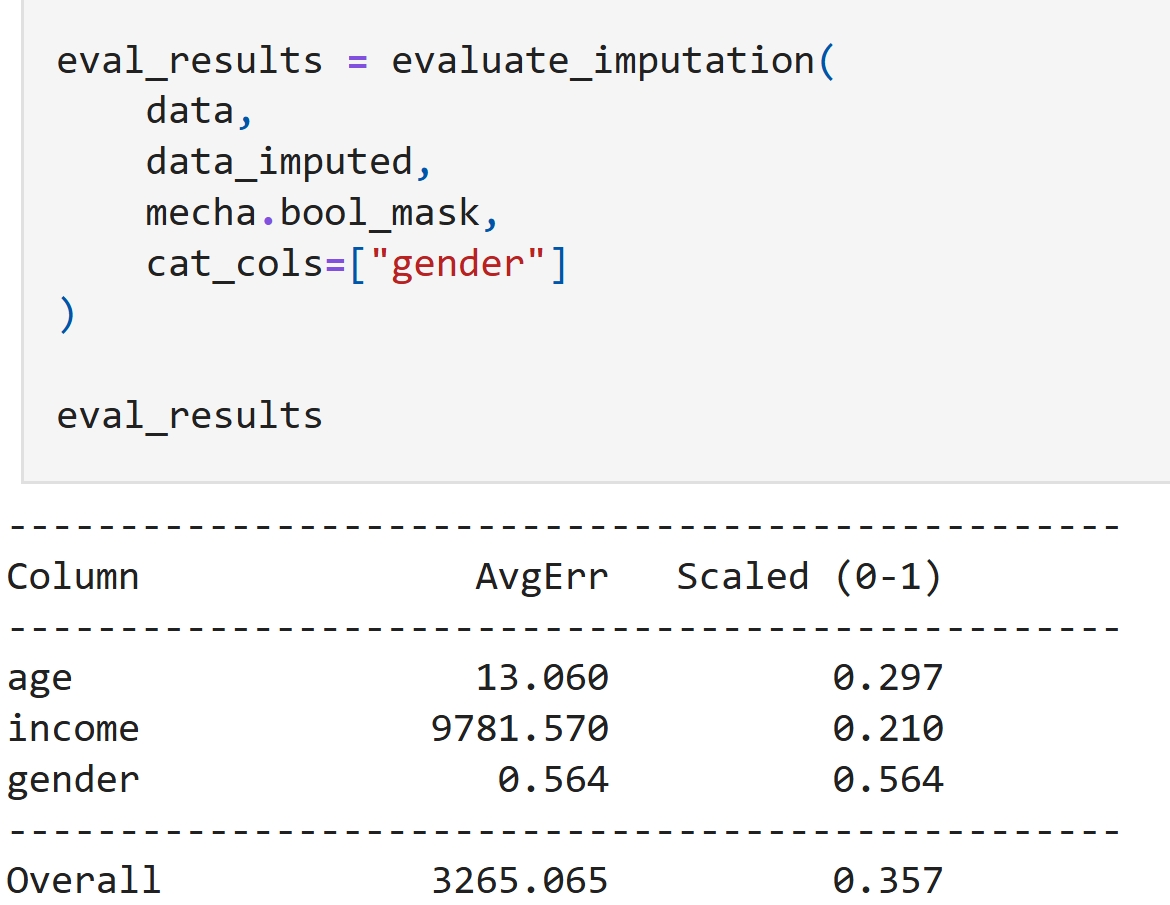}
    \caption{Evaluation report (\texttt{AvgErr}).}
  \end{subfigure}
  \hfill
  \begin{subfigure}{0.32\textwidth}
    \includegraphics[width=\linewidth]{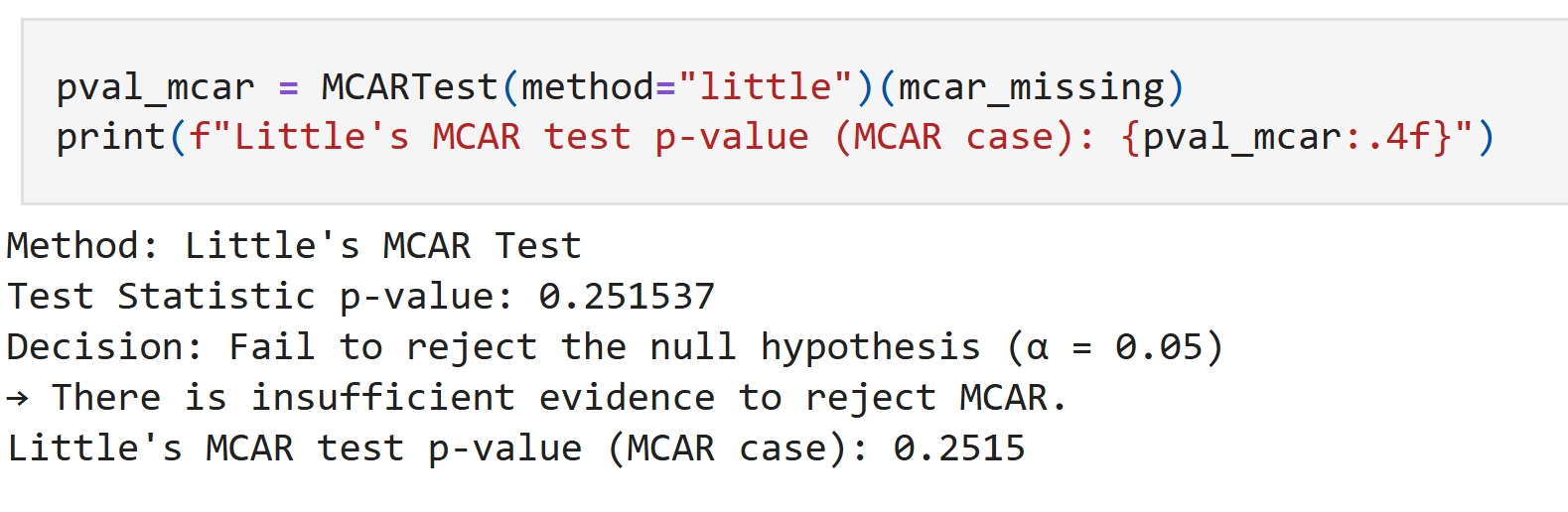}
    \caption{Little’s MCAR test output.}
  \end{subfigure}
  \hfill
  \begin{subfigure}{0.32\textwidth}
    \includegraphics[width=\linewidth]{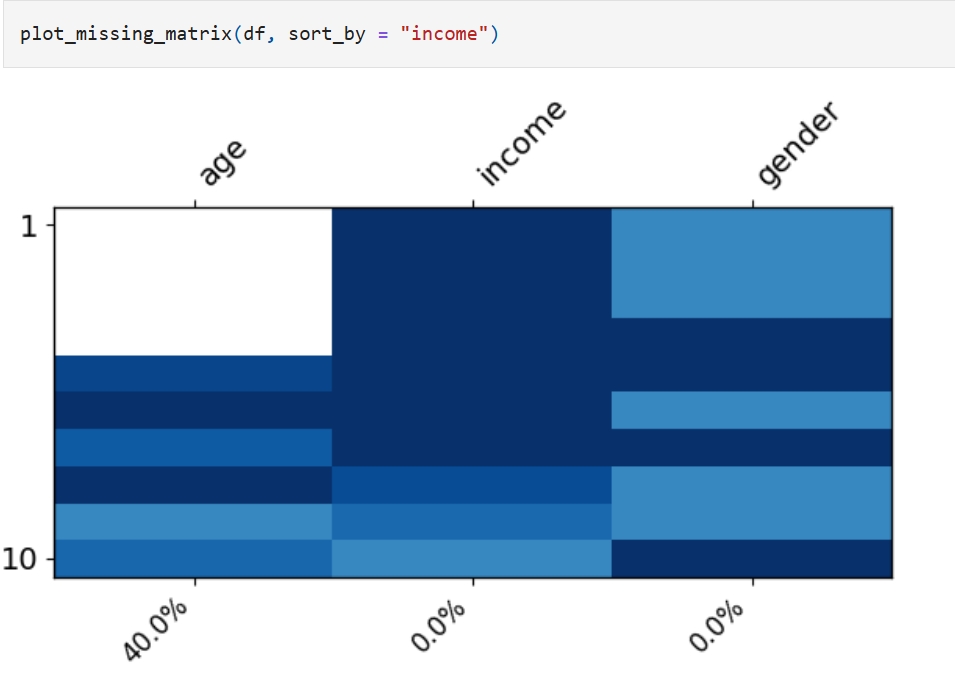}
    \caption{Missing Pattern Heatmap.}
  \end{subfigure}

  \caption{ The panels show (A) MCAR-based simulation; (B) per-feature missing rate summary; (C) imputer setup and; (D)type-aware evaluation; (E) MCAR testing; and (F) Heatmap to see missing pattern.}
  \label{fig:missmecha-demo}
\end{figure*}

\subsection{Visual Module}

The \texttt{visual} module extends \texttt{missingno}~\cite{missingno} with more customizable, scalable, and type-aware plots for inspecting missing data. It directly supports both numerical and categorical variables without preprocessing and scales well to large or time-indexed datasets. A key feature is nullity correlation analysis, which computes pairwise correlations between binary masks of missingness to detect potential MAR or MNAR patterns. Multiple methods are supported (\texttt{pearson}, \texttt{spearman}, \texttt{kendall}). The module provides two main functions: \texttt{plot\_missing\_matrix}, which shows a binary heatmap of missing entries with auto-scaled layout; and \texttt{plot\_missing\_heatmap}, which visualizes nullity dependencies without requiring one-hot encoding or normalization. All plots return \texttt{matplotlib} \texttt{Axes} objects and support full customization for figure aesthetics and integration into analysis pipelines.

\subsection{Impute Module}

The \texttt{impute} module offers a lightweight baseline imputer, \\\texttt{SimpleSmartImputer}, for quick evaluation and teaching. It automatically applies mean imputation for numerical columns and mode for categorical ones, with optional type inference or user-specified \texttt{cat\_cols}.The interface follows \texttt{scikit-learn} conventions with \texttt{fit}, \texttt{transform}, and \texttt{fit\_transform} methods. A verbose mode provides summaries of applied strategies, aiding interpretability. While not model-based, this imputer is well-suited for testing mechanism effects or initializing pipelines on mixed-type data.

\section{MissMecha Demonstration }

\paragraph{Scenario: Mechanism-Aware Imputation Analysis}
We demonstrate the core functionalities of \texttt{MissMecha} through a practical scenario where a data scientist explores how different missingness mechanisms impact imputation and diagnostic outcomes. \textbf{Setup.}  
Suppose you are analyzing a fully observed customer dataset. To evaluate the robustness of imputation strategies, you introduce structured missingness using \texttt{MissMechaGenerator}. Specifically, MAR-Type1 is applied to features, with a global missing rate of 50\%. \textbf{1. Simulate Missingness.}  
Panel (A) shows missing values generated under a standard MCAR setting using uniform masking. \textbf{2. Summarize Missing Rate.}  
Panel (B) presents the per-column missingness summary, useful for detecting sparse or skewed missing patterns. \textbf{3. Impute Missing Values.}  
Panel (C) shows the configuration of \texttt{SimpleSmartImputer}, which applies mean imputation for numerical features and mode for categorical features. \textbf{4. Evaluate Imputation Quality.}  Panel (D) displays the evaluation report using \texttt{AvgErr}, a hybrid metric that computes reconstruction errors separately by feature type. \textbf{5. Test for MCAR Assumption.}  
Panel (E) demonstrates how to use Little’s MCAR test to assess whether the missingness pattern could be considered completely at random.
\paragraph{Other Example: Visualize Missing Pattern Heatmap.}
Panel (F) displays a heatmap showing the relationship between missingness and feature values across columns. Darker regions may indicate that missing values tend to occur when certain columns take on specific ranges, helping users detect structured missingness.

\section{Conclusion and Future Work}
We introduced \texttt{MissMecha}, a Python toolkit for simulating, visualizing, and evaluating missing data mechanisms in mixed-type tabular data. It supports MCAR, MAR, and MNAR generation, type-aware evaluation, and statistical diagnostics. Future work will extend support to temporal missingness, advanced metrics, and customizable mechanisms, establishing \texttt{MissMecha} as a platform for reproducible and extensible missing data research.


\section*{GenAI Usage Disclosure}
We used GitHub Copilot and ChatGPT-4 for initial docstrings and code comments. All algorithm design, implementation, and writing were completed and reviewed by the authors. The demo video narration was generated using an AI voice, based on a script written and verified by the authors.

\bibliographystyle{ACM-Reference-Format}
\bibliography{sample-base}

\end{document}